%% file: acl2020.tex
\documentclass[11pt,a4paper]{article}
\usepackage[hyperref]{acl2020}
\usepackage{times}
\usepackage{latexsym}
\usepackage{amssymb}
\usepackage{amsmath}
\usepackage{booktabs}
\usepackage{multirow}
\usepackage{multicol}
\usepackage{tipa} 

\usepackage[]{todonotes}

\newcommand*\iftodonotes{\if@todonotes@disabled\expandafter\@secondoftwo\else\expandafter\@firstoftwo\fi}

\usepackage{microtype}

\aclfinalcopy 
\def\aclpaperid{280} 

\title{Adversarial NLI: A New Benchmark \\for Natural Language Understanding}

\author{Yixin Nie$^*$, Adina Williams$^\dagger$, Emily Dinan$^\dagger$, Mohit Bansal$^*$, Jason Weston$^\dagger$, Douwe Kiela$^\dagger$\\
  $^*$UNC Chapel Hill\\$^\dagger$Facebook AI Research\\}

\date{}

\begin{document}
\maketitle
\begin{abstract}
We introduce a new large-scale NLI benchmark dataset, collected via an iterative, adversarial human-and-model-in-the-loop procedure. We show that training models on this new dataset leads to state-of-the-art performance on a variety of popular NLI benchmarks, while posing a more difficult challenge with its new test set. Our analysis sheds light on the shortcomings of current state-of-the-art models, and shows that non-expert annotators are successful at finding their weaknesses. The data collection method can be applied in a never-ending learning scenario, becoming a moving target for NLU, rather than a static benchmark that will quickly saturate.
\end{abstract}

\section{Introduction}

Progress in AI has been driven by, among other things, the development of challenging large-scale  benchmarks like ImageNet~\cite{Russakovsky2015imagenet} in computer vision, and SNLI~\cite{Bowman2015snli}, SQuAD~\cite{Rajpurkar2016squad}, and others in natural language processing (NLP). Recently, for natural language understanding (NLU) in particular, the focus has shifted to combined benchmarks like SentEval~\cite{Conneau2018senteval} and GLUE~\cite{Wang2018glue}, which track model performance on multiple tasks and provide a unified platform for analysis. 

With the rapid pace of advancement in AI, however, NLU benchmarks struggle to keep up with model improvement. Whereas it took around 15 years to achieve ``near-human performance'' on MNIST~\cite{Lecun1998gradient,Cirecsan2012multi,Wan2013dropconnect} and approximately 7 years to surpass humans on ImageNet~\cite{Deng2009imagenet,Russakovsky2015imagenet,He2016resnet}, the GLUE benchmark did not last as long as we would have hoped after the advent of BERT~\cite{Devlin2018bert}, and rapidly had to be extended into SuperGLUE~\cite{Wang2019superglue}. This raises an important question: Can we collect a large benchmark dataset that can last longer?

The speed with which benchmarks become obsolete raises another important question: are current NLU models genuinely as good as their high performance on benchmarks suggests? A growing body of evidence shows that state-of-the-art models learn to exploit spurious statistical patterns in datasets~\cite{Gururangan2018annotation,Poliak2018hypothesis,Tsuchiya2018performance,Glockner2018breaking,Geva2019taskorannotator,Mccoy2019right}, instead of learning \emph{meaning} in the flexible and generalizable way that humans do. Given this, human annotators---be they seasoned NLP researchers or non-experts---might easily be able to construct examples that expose model brittleness.  %

\begin{figure*}[ht] 
	\centering
    \includegraphics[clip,width=0.95\textwidth]{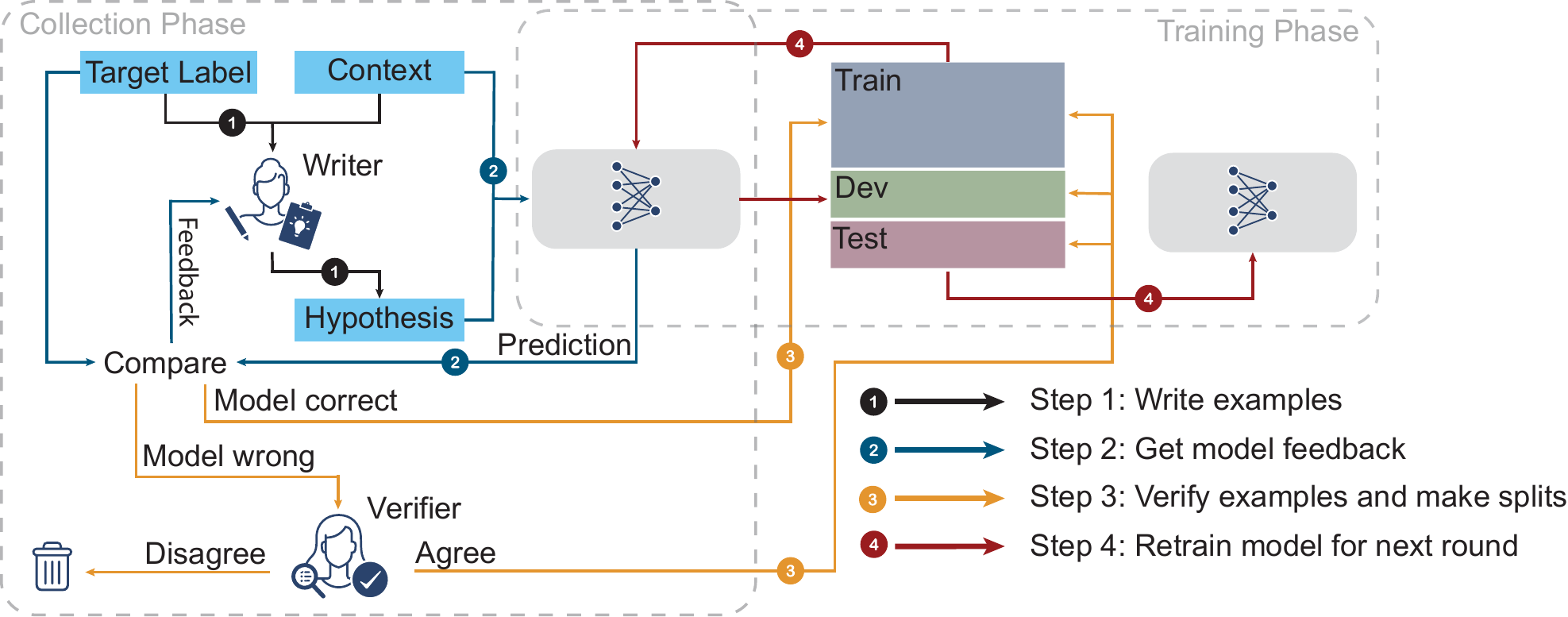}
\caption{Adversarial NLI data collection via human-and-model-in-the-loop enabled training~(HAMLET). The four steps make up one round of data collection. In step 3, model-correct examples are included in the training set; development and test sets are constructed solely from model-wrong verified-correct examples. \label{fig:process}}
\end{figure*}


We propose an iterative, adversarial human-and-model-in-the-loop solution for NLU dataset collection that addresses both benchmark longevity and robustness issues. In the first stage, human annotators devise examples that our current best models cannot determine the correct label for. These resulting hard examples---which should expose additional model weaknesses---can be added to the training set and used to train a stronger model. We then subject the strengthened model to the same procedure and collect weaknesses over several rounds. After each round, we train a new model and set aside a new test set. The process can be iteratively repeated in a never-ending learning \cite{Mitchell2018nell} setting
,
with the model getting stronger and the test set getting harder in each new round. %
Thus, not only is the resultant dataset harder than existing benchmarks, but this process also yields a ``moving post'' dynamic target for NLU systems, rather than a static benchmark that will eventually saturate. 

Our approach draws inspiration from recent efforts that gamify collaborative training of machine learning agents over multiple
rounds~\cite{Yang2017mastering} and pit ``builders'' against ``breakers'' to learn better models~\cite{Ettinger2017towards}. Recently, \newcite{Dinan2019build} showed that such an approach can be used to make dialogue safety classifiers more robust.
Here, we focus on natural language inference (NLI), arguably the most canonical task in NLU. We collected three rounds of data, and call our new dataset Adversarial NLI (ANLI).


Our contributions are as follows: 1) We introduce a novel human-and-model-in-the-loop dataset, consisting of three rounds that progressively increase in difficulty and complexity, that includes annotator-provided explanations. 2) We show that training models on this new dataset leads to state-of-the-art performance on a variety of popular NLI benchmarks. 3) We provide a detailed analysis of the collected data that sheds light on the shortcomings of current models, categorizes the data by inference type to examine weaknesses, and demonstrates good performance on NLI stress tests. The ANLI dataset is available at 
\href{https://github.com/facebookresearch/anli/}{github.com/facebookresearch/anli/}. A demo is available at \href{https://www.adversarialnli.com}{adversarialnli.com}.

\begin{table*}[ht]
\centering
\tiny
	\begin{tabular}{p{19em}p{9em}p{14em}p{4em}ccrp{7.7em}}
		\toprule
        \multirow{2}{*}{\bf Context} &  \multirow{2}{*}{\bf Hypothesis} &  \multirow{2}{*}{\bf Reason} &
        \multirow{2}{*}{\bf Round} & \multicolumn{3}{c}{\bf Labels} & \multirow{2}{*}{\bf Annotations} \\
         & & & & orig. & pred. & valid. & \\
		\midrule
	Roberto Javier Mora Garc\'ia (c. 1962 -- 16 March 2004) was a Mexican journalist and editorial director of ``El Ma\~nana'', a newspaper based in Nuevo Laredo, Tamaulipas, Mexico. He worked for a number of media outlets in Mexico, including the ``El Norte'' and ``El Diario de Monterrey'', prior to his assassination. & Another individual laid waste to Roberto Javier Mora Garcia. & The context states that Roberto Javier Mora Garcia was assassinated, so another person had to have ``laid waste to him.'' The system most likely had a hard time figuring this out due to it not recognizing the phrase ``laid waste.'' & A1 (Wiki) & E  & N & E E & Lexical (assassination, laid waste), Tricky (Presupposition), Standard (Idiom)\\ \midrule
	 A melee weapon is any weapon used in direct hand-to-hand combat; by contrast with ranged weapons which act at a distance. The term ``melee'' originates in the 1640s from the French word ``m\u{e}l\'ee'', which refers to hand-to-hand combat, a close quarters battle, a brawl, a confused fight, etc. Melee weapons can be broadly divided into three categories & Melee weapons are good for ranged and hand-to-hand combat. & Melee weapons are good for hand to hand combat, but NOT ranged. & A2 (Wiki) & C & E & C N C & Standard (Conjunction), Tricky (Exhaustification), Reasoning (Facts)\\ \midrule
	 If you can dream it, you can achieve it---unless you're a goose trying to play a very human game of rugby. In the video above, one bold bird took a chance when it ran onto a rugby field mid-play. Things got dicey when it got into a tussle with another player, but it shook it off and kept right on running. After the play ended, the players escorted the feisty goose off the pitch. It was a risky move, but the crowd chanting its name was well worth it. & The crowd believed they knew the name of the goose running on the field. & Because the crowd was chanting its name, the crowd must have believed they knew the goose's name. The word ``believe'' may have made the system think this was an ambiguous statement. & A3 (News) & E & N & E E & Reasoning (Facts), Reference (Coreference) \\
		\bottomrule
	\end{tabular}
	\caption{\label{tab:examples} Examples from development set. `A$n$' refers to round number, `orig.' is the original annotator's gold label, `pred.' is the model prediction, `valid.' are the validator labels, `reason' was provided by the original annotator, `Annotations' are the tags determined by an linguist expert annotator.
	} 
\end{table*}

\section{Dataset collection}

The primary aim of this work is to create a new large-scale NLI benchmark on which current state-of-the-art models fail. This constitutes a new target for the field to work towards, and can elucidate model capabilities and limitations. As noted, however, static benchmarks do not last very long these days. If continuously deployed, the data collection procedure we introduce here can pose a dynamic challenge that allows for never-ending learning.

\subsection{HAMLET}
To paraphrase the great bard \cite{Shakespeare1603hamlet}, \emph{there is something rotten in the state of the art}. We propose \emph{Human-And-Model-in-the-Loop Enabled Training} (HAMLET), a training procedure to automatically mitigate problems with current dataset collection procedures (see Figure \ref{fig:process}).

In our setup, our starting point is a \emph{base model}, trained on NLI data. Rather than employing automated adversarial methods, here the model's ``adversary'' is a human annotator. Given a \emph{context} (also often called a ``premise'' in NLI), and a desired \emph{target label}, we ask the human \emph{writer} to provide a \emph{hypothesis} that fools the model into misclassifying the label. One can think of the writer as a ``white hat'' hacker, trying to identify vulnerabilities in the system
. For each human-generated example that is misclassified, we also ask the writer to provide a \emph{reason} why they believe it was misclassified.

For examples that the model misclassified, it is necessary to verify that they are actually correct
---i.e., that the given context-hypothesis pairs genuinely have their specified target label. The best way to do this is to have them checked by another human. Hence, we provide the example to human \emph{verifiers}. If two human verifiers agree with the writer, the example is considered a good example. If they disagree, we ask a third human verifier to break the tie. If there is still disagreement between the writer and the verifiers, the example is discarded. If the verifiers disagree, they can overrule the original target label of the writer.

Once data collection for the current round is finished, we construct a new training set from the collected data, with accompanying development and test sets, which are constructed solely from verified correct examples. The test set was further restricted so as to: 1) include pairs from ``exclusive'' annotators who are never included in the training data; and 2) be balanced by label classes (and genres, where applicable). We subsequently train a \emph{new model} on this and other existing data, and repeat the procedure. 

\subsection{Annotation details}

We employed Mechanical Turk workers with qualifications and collected hypotheses via the ParlAI\footnote{\url{https://parl.ai/}} framework.
Annotators are presented with a context and a target label---either `entailment', `contradiction', or `neutral'---and asked to write a hypothesis that corresponds to the label. We phrase the label classes as ``definitely correct'', ``definitely incorrect'', or ``neither definitely correct nor definitely incorrect'' given the context, to make the task easier to grasp. 
Model predictions are obtained for the context and submitted hypothesis pair. 
The probability of each label is shown to the worker as feedback. If the model prediction was incorrect, the job is complete. If not, the worker continues to write hypotheses for the given (context, target-label) pair until the model predicts the label incorrectly or the number of tries exceeds a threshold (5 tries in the first round, 10 tries thereafter). 

To encourage workers, payments increased as rounds became harder. For hypotheses that the model predicted incorrectly, and that were verified by other humans, we paid an additional bonus on top of the standard rate.


\subsection{Round 1}

For the first round, we used a BERT-Large model \cite{Devlin2018bert} trained on a concatenation of SNLI \cite{Bowman2015snli} and MNLI \cite{Williams2017mnli}, and selected the best-performing model we could train as the starting point for our dataset collection procedure. For Round 1 contexts, we randomly sampled short multi-sentence passages from Wikipedia~(of 250-600 characters)~from the manually curated HotpotQA training set \cite{Yang2018hotpotqa}. 
Contexts are either ground-truth contexts 
from that dataset, or they are Wikipedia passages retrieved using TF-IDF \cite{chen2017drqa} based on a HotpotQA question.

\begin{table*}[t]
    \centering
    \small
    \begin{tabular}{lccrrrrr} 
    \toprule
    \textbf{Dataset} & \textbf{Genre} & \textbf{Context} & \textbf{Train / Dev / Test} & \multicolumn{2}{c}{\textbf{Model error rate}}
    & \textbf{Tries} & \textbf{Time (sec.)}
    \\
    & & & & Unverified & Verified & \multicolumn{2}{c}{\scriptsize mean/median per verified ex.}\\
    \midrule
    A1 & Wiki & 2,080 & 16,946 / 1,000 / 1,000 & 29.68\% & 18.33\% & 3.4 / 2.0  & 199.2 / 125.2 \\ \midrule
    A2 & Wiki & 2,694 & 45,460 / 1,000 / 1,000 & 16.59\% & 8.07\% & 6.4 / 4.0 & 355.3 / 189.1 \\\midrule
    \multirow{2}{*}{A3} & Various & 6,002 & 100,459 / 1,200 / 1,200 & 17.47\% & 8.60\% & 6.4 / 4.0 & 284.0 / 157.0 \\
     & (Wiki subset) & 1,000 & 19,920 / 200 / 200 & 14.79\% & 6.92\% & 7.4 / 5.0 & 337.3 / 189.6 \\\midrule
    ANLI & Various & 10,776 & 162,865 / 3,200 / 3,200 & 18.54\% & 9.52\%  & 5.7 / 3.0 & 282.9 / 156.3\\
    \bottomrule
    \end{tabular}
    \caption{Dataset statistics: `Model error rate' is the percentage of examples that the model got wrong; `unverified' is the overall percentage, while `verified' is the percentage that was verified by at least 2 human annotators.}
    \label{tab:dataset_statistics}
\end{table*}

\subsection{Round 2}

For the second round, we used a more powerful RoBERTa model \cite{Liu2019roberta} trained on SNLI, MNLI, an NLI-version\footnote{The NLI version of FEVER pairs claims with evidence retrieved by \newcite{nie2019combining} as (context, hypothesis) inputs.} of FEVER \cite{Thorne2018fever}, and the training data from the previous round (A1). After a hyperparameter search, we selected the model with the best performance on the A1 development set.
Then, using the hyperparameters selected from this search, we created a final set of models by training several models with different random seeds. During annotation, we constructed an ensemble by randomly picking a model from the model set as the adversary each turn. This helps us avoid annotators exploiting vulnerabilities in one single model. A new non-overlapping set of contexts was again constructed from Wikipedia via HotpotQA using the same method as Round 1.

\subsection{Round 3}

For the third round, we selected a more diverse set of contexts, in order to explore robustness under domain transfer. In addition to contexts from Wikipedia for Round 3, we also included contexts from the following domains: News (extracted from Common Crawl), fiction (extracted from StoryCloze \cite{Mostafazadeh2016storycloze} and CBT \cite{Hill2015goldilocks}), formal spoken text (excerpted from court and presidential debate transcripts in the Manually Annotated Sub-Corpus (MASC) of the Open American National Corpus\footnote{\url{anc.org/data/masc/corpus/}}), and causal or procedural text, which describes sequences of events or actions, extracted from WikiHow. Finally, we also collected annotations using the longer contexts present in the GLUE RTE training data, which came from the RTE5 dataset \cite{Bentivogli2009rte5}. We trained an even stronger RoBERTa ensemble by adding the training set from the second round (A2) to the training data.

\subsection{Comparing with other datasets}

The ANLI dataset, comprising three rounds, improves upon previous work in several ways. First, and most obviously, the dataset is collected to be more difficult than previous datasets, by design. Second, it remedies a problem with SNLI, namely that its contexts (or premises) are very short, because they were selected from the image captioning domain. We believe longer contexts should naturally lead to harder examples, and so we constructed ANLI contexts from longer, multi-sentence source material.

Following previous observations that models might exploit spurious biases in NLI hypotheses, \cite{Gururangan2018annotation,Poliak2018hypothesis}, we conduct a study of the performance of hypothesis-only models on our dataset. We show that such models perform poorly on our test sets.

With respect to data generation with na\"ive annotators, \newcite{Geva2019taskorannotator} noted that models can pick up on annotator bias, modelling annotator artefacts rather than the intended reasoning phenomenon. To counter this, we selected a subset of annotators (i.e., the ``exclusive'' workers) whose data would only be included in the test set. This enables us to avoid overfitting to the writing style biases of particular annotators, and also to determine how much individual annotator bias is present for the main portion of the data. Examples from each round of dataset collection are provided in Table \ref{tab:examples}.

Furthermore, our dataset poses new challenges to the community that were less relevant for previous work, such as: can we improve performance online without having to train a new model from scratch every round, how can we overcome catastrophic forgetting, how do we deal with mixed model biases, etc. Because the training set includes examples that the model got right but were not verified, learning from noisy and potentially unverified data becomes an additional interesting challenge.

\begin{table*}[t]
    \centering
    \small
    \begin{tabular}{llrrrrr|rr}
    \toprule
    \textbf{Model} & \textbf{Training Data} & A1 & A2 & A3 & ANLI & ANLI-E & SNLI & MNLI-m/-mm\\ \midrule
 \multirow{5}{*}{BERT} & S,M\textbf{$^{\star 1}$} & \underline{00.0} & 28.9 & 28.8 & 19.8 & 19.9 & 91.3 & 86.7 / 86.4 \\
 	& +A1 & 44.2 & 32.6 & 29.3 & 35.0 & 34.2 & 91.3 & 86.3 / 86.5 \\
 	& +A1+A2 & 57.3 & 45.2 & 33.4 & 44.6 & 43.2 & 90.9 & 86.3 / 86.3\\
 	& +A1+A2+A3 & 57.2 & 49.0 & 46.1 & 50.5 & 46.3 & 90.9 & 85.6 / 85.4\\
 	& S,M,F,ANLI & 57.4 & 48.3 & 43.5 & 49.3 & 44.2 & 90.4 & 86.0 / 85.8\\\midrule
 XLNet & S,M,F,ANLI & 67.6 & 50.7 & 48.3 & 55.1 &  52.0 & 91.8 & 89.6 / 89.4 \\\midrule 
 \multirow{5}{*}{RoBERTa} & S,M & 47.6 & 25.4 & 22.1 & 31.1 & 31.4 & 92.6 & 90.8 / 90.6 \\
 	& +F & 54.0 & 24.2 & 22.4 & 32.8 & 33.7 & 92.7 & 90.6 / 90.5 \\
 	& +F+A1\textbf{$^{\star 2}$} & 68.7 & \underline{19.3} & 22.0 & 35.8 & 36.8 & 92.8 & 90.9 / 90.7\\
 	& +F+A1+A2\textbf{$^{\star 3}$} & 71.2 & 44.3 & \underline{20.4} & 43.7
 	& 41.4 & 92.9 & 91.0 / 90.7 \\
 	& S,M,F,ANLI & 73.8 & 48.9 & 44.4 & 53.7 & 49.7 & 92.6 & 91.0 / 90.6 \\\bottomrule
     \end{tabular}
     \vspace{-5pt}
     \caption{Model Performance. `S' refers to \textsc{SNLI}, `M' to \textsc{MNLI} dev (-m=matched, -mm=mismatched), and `F' to \textsc{FEVER}; `A1--A3' refer to the rounds respectively and `ANLI' refers to A1+A2+A3, `-E' refers to test set examples written by annotators exclusive to the test set.
     Datasets marked `$^{\star n}$' were used to train the base model for round $n$, and their performance on that round is \underline{underlined} (A2 and A3 used ensembles, and hence have non-zero scores).}
     \label{tab:model_performance}
     \vspace{-8pt}
\end{table*}

\section{Dataset statistics}

The dataset statistics can be found in Table \ref{tab:dataset_statistics}. The number of examples we collected increases per round, starting with approximately 19k examples for Round 1, to around 47k examples for Round 2, to over 103k examples for Round 3. We collected more data for later rounds not only because that data is likely to be more interesting, but also simply because the base model is better and so annotation took longer to collect good, verified correct examples of model vulnerabilities.

For each round, we report the model error rate, both on verified and unverified examples. The unverified model error rate captures the percentage of examples where the model disagreed with the writer's target label, but where we are not (yet) sure if the example is correct. The verified model error rate is the percentage of model errors from example pairs that other annotators confirmed the correct label for. Note that error rate is a useful way to evaluate model quality: the lower the model error rate---assuming constant annotator quality and context-difficulty---the better the model.

We observe that model error rates decrease as we progress through rounds. In Round 3, where we included a more diverse range of contexts from various domains, the overall error rate went slightly up compared to the preceding round, but for Wikipedia contexts the error rate decreased substantially. While for the first round roughly 1 in every 5 examples were verified model errors, this quickly dropped over consecutive rounds, and the overall model error rate is less than 1 in 10. On the one hand, this is impressive, and shows how far we have come with just three rounds. On the other hand, it shows that we still have a long way to go if even untrained annotators can fool ensembles of state-of-the-art models with relative ease.

Table \ref{tab:dataset_statistics} also reports the average number of ``tries'', i.e., attempts made for each context until a model error was found (or the number of possible tries is exceeded), and the average time this took (in seconds). Again, these metrics are useful for evaluating model quality: observe that the average number of tries and average time per verified error both go up with later rounds. This demonstrates that the rounds are getting increasingly more difficult. Further dataset statistics and inter-annotator agreement are reported in Appendix \ref{appendix:datasetproperties}.

\section{Results}

%

%
Table~\ref{tab:model_performance} reports the main results. In addition to BERT \cite{Devlin2018bert} and RoBERTa \cite{Liu2019roberta}, we also include XLNet \cite{Yang2019xlnet} as an example of a strong, but different, model architecture. We show test set performance on the ANLI test sets per round, the total ANLI test set, and the exclusive test subset (examples from test-set-exclusive workers). We also show accuracy on the SNLI test set and the MNLI development set (for the purpose of comparing between different model configurations across table rows). In what follows, we discuss our observations.

\paragraph{Base model performance is low.} Notice that the base model for each round performs very poorly on that round's test set. This is the expected outcome: For round 1, the base model gets the entire test set wrong, by design. For rounds 2 and 3, we used an ensemble, so performance is not necessarily zero. However, as it turns out, performance still falls well below chance\footnote{Chance is at 33\%, since the test set labels are balanced.}, indicating that workers did not find vulnerabilities specific to a single model, but generally applicable ones for that model class.

\paragraph{Rounds become increasingly more difficult.} As already foreshadowed by the dataset statistics, round 3 is more difficult (yields lower performance) than round 2, and round 2 is more difficult than round 1. This is true for all model architectures.

\paragraph{Training on more rounds improves robustness.} 
Generally, our results indicate that training on more rounds improves model performance. This is true for all model architectures. Simply training on more ``normal NLI'' data would not help a model be robust to adversarial attacks, but our data actively helps mitigate these.

\paragraph{RoBERTa achieves state-of-the-art performance...} We obtain state of the art performance on both SNLI and MNLI with the RoBERTa model finetuned on our new data. The RoBERTa paper~\cite{Liu2019roberta} reports a score of $90.2$ for both MNLI-matched and -mismatched dev, while we obtain $91.0$ and $90.7$. The state of the art on SNLI is currently held by MT-DNN~\cite{Liu2019mtdnn}, which reports $91.6$ compared to our $92.9$.

\paragraph{...but is outperformed when it is base model.} However, the base (RoBERTa) models for rounds 2 and 3 are outperformed by both BERT and XLNet (rows 5, 6 and 10). This shows that annotators found examples that RoBERTa generally struggles with, which cannot be mitigated by more examples alone. It also implies that BERT, XLNet, and RoBERTa all have different weaknesses, possibly as a function of their training data (BERT, XLNet and RoBERTa were trained on different data sets, which might or might not have contained information relevant to the weaknesses).

\begin{figure}[t]
    \centering
    \includegraphics[width=.45\textwidth]{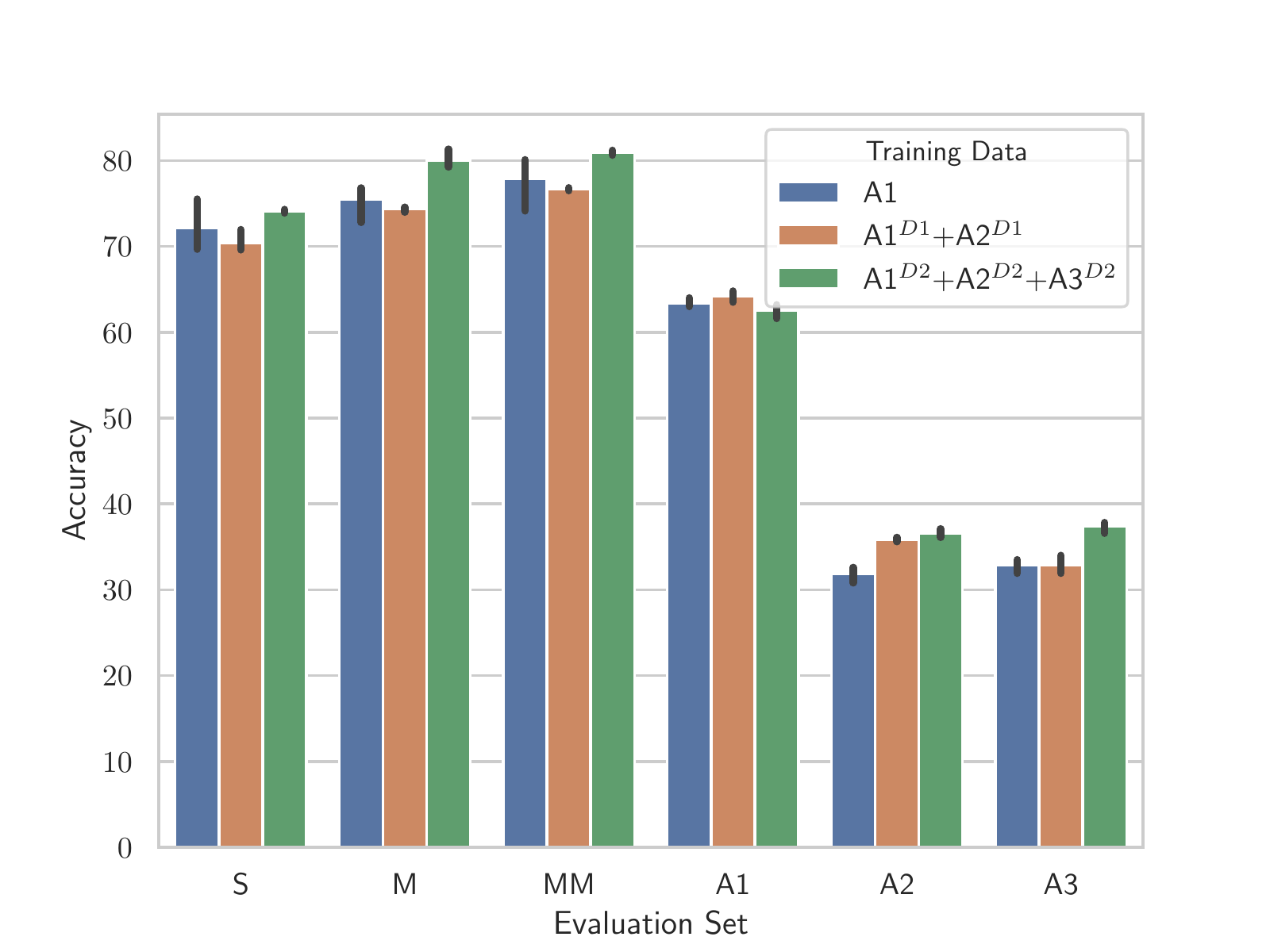}
    \vspace{-5pt}
    \caption{RoBERTa performance on dev, with A1--3 downsampled s.t. $|$A1$^{D1}|$=$|$A2$^{D1}|$=$\frac{1}{2}|$A1$|$ and $|$A1$^{D2}|$=$|$A2$^{D2}|$=$|$A3$^{D2}|$=$\frac{1}{3}|$A1$|$.}
    \label{fig:data-efficiency}
\vspace{-10pt}
\end{figure}

\paragraph{Continuously augmenting training data does not downgrade performance.} Even though ANLI training data is different from SNLI and MNLI, adding it to the training set does not harm performance on those tasks. Our results (see also rows 2-3 of Table~\ref{tab:HO}) suggest the method could successfully be applied for multiple additional rounds.


\paragraph{Exclusive test subset difference is small.} 
We included an exclusive test subset (ANLI-E) with examples from annotators never seen in training, and find negligible differences, indicating that our models do not over-rely on annotator's writing styles.

\begin{figure}[t]
    \centering
    \includegraphics[width=.45\textwidth]{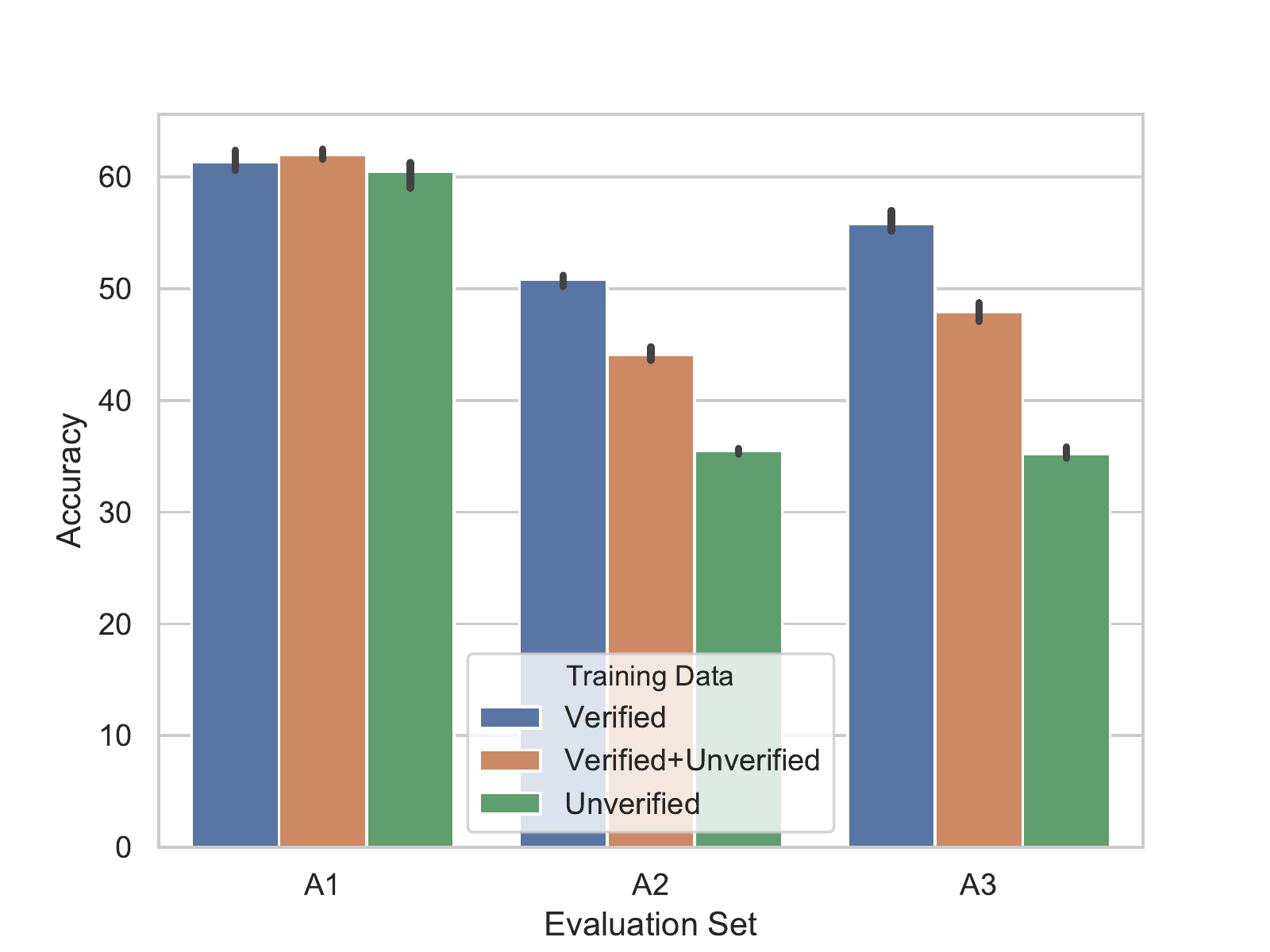}
    \vspace{-5pt}
    \caption{Comparison of verified, unverified and combined data, where data sets are downsampled to ensure equal training sizes.}
    \label{fig:verified-v-unverified}
\vspace{-15pt}
\end{figure}

\subsection{The effectiveness of adversarial training}

We examine the effectiveness of the adversarial training data in two ways. First, we sample from respective datasets to ensure exactly equal amounts of training data. Table \ref{tab:Downsample_performance} shows that the adversarial data improves performance, including on SNLI and MNLI when we replace part of those datasets with the adversarial data. This suggests that the adversarial data is more data efficient than ``normally collected'' data. Figure \ref{fig:data-efficiency} shows that adversarial data collected in later rounds is of higher quality and more data-efficient.

\begin{table*}[ht]
    \centering
    \small
    \begin{tabular}{lrrrrrrr}
    \toprule
    \multirow{2}{*}{\textbf{Model}} & \multirow{2}{*}{\textbf{SNLI-Hard}} & \multicolumn{6}{c}{\textbf{NLI Stress Tests}} \\
    \cmidrule(lr){3-8}
     & & \small{AT (m/mm)} & NR & LN (m/mm) & NG (m/mm) & WO (m/mm) & SE (m/mm)\\
    \midrule
    Previous models & 72.7 & 14.4 / 10.2 & 28.8 & 58.7 / 59.4 & 48.8 / 46.6 & 50.0 / 50.2 & 58.3 / 59.4 \\\midrule
    BERT (All) & 82.3 & 75.0 / 72.9 & 65.8 & 84.2 / 84.6 & 64.9 / 64.4 & 61.6 / 60.6 & 78.3 / 78.3	\\
    XLNet (All) & 83.5 & 88.2	/ 87.1 &	85.4 & 87.5 / 87.5 & 59.9 / 60.0 & 	68.7 / 66.1 & 84.3 / 84.4	\\
    RoBERTa (S+M+F) & 84.5 & 81.6	/ 77.2 &	62.1 & 88.0 / 88.5 & 61.9 / 61.9 &	67.9 / 66.2 & 86.2 / 86.5	
    \\
    RoBERTa (All) & 84.7 & 85.9 / 82.1 &	80.6 & 88.4 / 88.5 & 62.2 / 61.9 &	67.4 / 65.6 & 86.3 / 86.7 \\
    \bottomrule
    \end{tabular}
    \vspace{-5pt}
    \caption{Model Performance on NLI stress tests (tuned on their respective dev. sets). All=S+M+F+ANLI. AT=`Antonym'; `NR'=Numerical Reasoning; `LN'=Length; `NG'=Negation; `WO'=Word Overlap; `SE'=Spell Error. Previous models refers to the \newcite{naik-EtAl:2018:C18-1} implementation of \newcite[InferSent]{conneau2017} for the Stress Tests, and to the \newcite{Gururangan2018annotation} implementation of \newcite[DIIN]{gong2018} for SNLI-Hard. }
    \label{tab:model_performance_prev_stress}
    \vspace{-5pt}
\end{table*}

\begin{table}[t]
    \centering
    \small 
    \begin{tabular}{lrrrrr}
        \toprule
        \textbf{Training Data} & A1 & A2 & A3 & S & M-m/mm\\\midrule
        SM$^{D1}$+SM$^{D2}$ & 45.1 & 26.1 & 27.1 & \textbf{92.5} & 89.8/\textbf{89.7}\\
        SM$^{D1}$+A & \textbf{72.6} & \textbf{42.9} & \textbf{42.0} & 92.3 & \textbf{90.3}/89.6\\
        \midrule
        SM & 48.0 & 24.8 & 31.1 & 93.2 & 90.8/90.6\\
        SM$^{D3}$+A & \textbf{73.3} & \textbf{42.4} & \textbf{40.5} & \textbf{93.3} & \textbf{90.8}/\textbf{90.7}\\\bottomrule
    \end{tabular}
    \vspace{-5pt}
    \caption{\label{tab:Downsample_performance} RoBERTa performance on dev set with different training data. S=SNLI, M=MNLI, A=A1+A2+A3. `SM' refers to combined S and M training set. D$1$, D$2$, D$3$ means down-sampling SM s.t. $|$SM$^{D2}|$=$|$A$|$ and $|$SM$^{D3}|$+$|$A$|$=$|$SM$|$. Therefore, training sizes are identical in every pair of rows.}
    \vspace{-8pt}
\end{table}

Second, we compared verified correct examples of model vulnerabilities (examples that the model got wrong and were verified to be correct) to unverified ones. Figure \ref{fig:verified-v-unverified} shows that the verified correct examples are much more valuable than the unverified examples, especially in the later rounds (where the latter drops to random). 

\subsection{Stress Test Results}
We also test models on two recent hard NLI test sets: SNLI-Hard \cite{Gururangan2018annotation} and the NLI stress tests \citep{naik-EtAl:2018:C18-1} (see Appendix~\ref{appendix:stresstests} for details). The results are in Table~\ref{tab:model_performance_prev_stress}. We observe that all our models outperform the models presented in original papers for these common stress tests.
The RoBERTa models perform best on SNLI-Hard and achieve accuracy levels in the high 80s on the `antonym' (AT), `numerical reasoning' (NR), `length' (LN), `spelling error'(SE) sub-datasets, and show marked improvement on both `negation' (NG), and `word overlap' (WO). Training on ANLI appears to be particularly useful for the AT, NR, NG and WO stress tests.


\subsection{Hypothesis-only results}\label{subsec:HO}

For SNLI and MNLI, concerns have been raised about the propensity of models to pick up on spurious artifacts that are present just in the hypotheses \cite{Gururangan2018annotation,Poliak2018hypothesis}. Here, we compare full models to models trained only on the hypothesis (marked $H$). Table \ref{tab:HO} reports results on ANLI, as well as on SNLI and MNLI. The table shows that hypothesis-only models perform poorly on ANLI\footnote{Obviously, without manual intervention, some bias remains in how people phrase hypotheses---e.g., contradiction might have more negation---which explains why hypothesis-only performs slightly above chance when trained on ANLI.}, and obtain good performance on SNLI and MNLI. Hypothesis-only performance decreases over rounds for ANLI.

\begin{table}[t]
    \centering
    \small 
    \begin{tabular}{lrrrrr}
        \toprule
        \textbf{Training Data} & A1 & A2 & A3 &
        S & M-m/mm\\\midrule
        ALL & 73.8 & 48.9 & 44.4 & 92.6 & 91.0/90.6\\
        S+M & 47.6 & 25.4 & 22.1 & 92.6 & 90.8/90.6\\
        ANLI-Only & 71.3 & 43.3 & 43.0 & 83.5 & 86.3/86.5 \\\midrule
        ALL$^{H}$ & 49.7 & 46.3 & 42.8 & 71.4 & 60.2/59.8\\
        S+M$^{H}$ & 33.1 & 29.4 & 32.2 & 71.8 & 62.0/62.0\\
        ANLI-Only$^{H}$ & 51.0 & 42.6 & 41.5 & 47.0 & 51.9/54.5\\
        \bottomrule
    \end{tabular}
    \vspace{-5pt}
    \caption{\label{tab:HO}
    Performance of RoBERTa with different data combinations. ALL=S,M,F,ANLI. Hypothesis-only models are marked $H$ where they are trained and tested with only hypothesis texts.}
    \vspace{-8pt}
\end{table}

We observe that in rounds 2 and 3, RoBERTa is not much better than hypothesis-only. This could mean two things: either the test data is very difficult, or the training data is not good. To rule out the latter, we trained only on ANLI ($\sim$163k training examples): RoBERTa matches BERT when trained on the much larger, fully in-domain SNLI+MNLI combined dataset (943k training examples) on MNLI, with both getting $\sim$86 (the third row in Table~\ref{tab:HO}). Hence, this shows that the test sets are so difficult that state-of-the-art models cannot outperform a hypothesis-only prior.

\begin{table*}[t]
    \centering
    \small
    \begin{tabular}{cccccccc}
    \toprule
    \textbf{Round} & Numerical \& Quant. & Reference \& Names & Standard & Lexical & Tricky & Reasoning \& Facts & Quality\\
    \midrule
    A1 & 38\% & 13\% & 18\% & 13\% & 22\% & 53\% & 4\% \\
    A2 & 32\% & 20\% & 21\% & 21\% & 20\% & 59\% & 3\% \\
    A3 & 10\% & 18\% & 27\% & 27\% &27\% & 63\% & 3\% \\\midrule
    Average & 27\%  & 17\% &  22\% & 22\% & 23\% & 58\% & 3\%  \\
    \bottomrule
    \end{tabular}
    \vspace{-5pt}
    \caption{Analysis of 500 development set examples per round and on average.
    }
    \label{tab:analysis}
    \vspace{-10pt}
\end{table*}

\section{Linguistic analysis}

We explore the types of inferences that fooled models by manually annotating $500$ examples from each round's development set. A dynamically evolving dataset offers the unique opportunity to track how model error rates change over time. Since each round's development set contains only verified examples, we can investigate two interesting questions: which types of inference do writers employ to fool the models, and are base models differentially sensitive to different types of reasoning?


The results  are summarized in Table \ref{tab:analysis}. We devised an inference ontology containing six types of inference: Numerical \& Quantitative (i.e., reasoning about cardinal and ordinal numbers, inferring dates and ages from numbers, etc.), Reference \& Names (coreferences between pronouns and forms of proper names, knowing facts about name gender, etc.), Standard Inferences (conjunctions, negations, cause-and-effect, comparatives and superlatives etc.), Lexical Inference (inferences made possible by lexical information about synonyms, antonyms, etc.), Tricky Inferences (wordplay, linguistic strategies such as syntactic transformations/reorderings, or inferring writer intentions from contexts), and reasoning from outside knowledge or additional facts (e.g., ``You can't reach the sea directly from Rwanda''). The quality of annotations was also tracked; if a pair was ambiguous or a label debatable (from the expert annotator's perspective), it was flagged. Quality issues were rare at 3-4\% per round. Any one example can have multiple types, and every example had at least one tag.



We observe that both round 1 and 2 writers rely heavily on numerical and quantitative reasoning  in over 30\% of the development set---the percentage in A2 (32\%) dropped roughly 6\% from A1 (38\%)---while round 3 writers use numerical or quantitative reasoning for only 17\%. The majority of numerical reasoning types were references to cardinal numbers that referred to dates and ages. Inferences predicated on references and names were present in about 10\% of rounds 1 \& 3 development sets, and reached a high of 20\% in round 2, with coreference featuring prominently. Standard inference types increased in prevalence as the rounds increased, ranging from 18\%--27\%, as did `Lexical' inferences (increasing from 13\%--31\%). The percentage of sentences relying on reasoning and outside facts remains roughly the same, in the mid-50s, perhaps slightly increasing over the rounds.  For round 3, we observe that the model used to collect it appears to be more susceptible to Standard, Lexical, and Tricky inference types. This finding is compatible with the idea that models trained on adversarial data perform better, since annotators seem to have been encouraged to devise more creative examples containing harder types of inference in order to stump them. Further analysis is provided in Appendix \ref{appendix:linguisticanalysis}.

\section{Related work}

\paragraph{Bias in datasets} Machine learning methods are well-known to pick up on spurious statistical patterns.
For instance,
in the first visual question answering dataset \cite{Antol2015vqa}, biases like ``2'' being the correct answer to 39\% of the questions starting with ``how many'' allowed learning algorithms to perform well while ignoring the visual modality altogether \cite{Jabri2016revisiting,Goyal2017makingvmatter}. 
In NLI, \newcite{Gururangan2018annotation},  \newcite{Poliak2018hypothesis} and \newcite{Tsuchiya2018performance} showed that hypothesis-only baselines often perform far better than chance. NLI systems can often be broken merely by performing simple lexical substitutions \cite{Glockner2018breaking}, and struggle with quantifiers \cite{Geiger2018stress} and certain superficial syntactic properties
\cite{Mccoy2019right}.

In question answering, \newcite{Kaushik2018howmuch} showed that question- and passage-only models can perform surprisingly well, while \newcite{Jia2017adversarial} added adversarially constructed sentences to passages to cause a drastic drop in performance. Many tasks do not actually require sophisticated linguistic reasoning, as shown by the surprisingly good performance of random encoders \cite{Wieting2019random}. Similar observations were made in machine translation \cite{Belinkov2017synthetic} and dialogue \cite{Sankar2019dialog}. Machine learning also has a tendency to overfit on static targets, even if that does not happen deliberately \cite{Recht2018cifar}. In short, the field is rife with dataset bias and papers trying to address this important problem. This work presents a potential solution: if such biases exist, they will allow humans to fool the models, resulting in valuable training examples until the bias is mitigated.

\paragraph{Dynamic datasets.}

\newcite{Bras2020aflite} proposed AFLite, an approach for avoiding spurious biases through adversarial filtering, which is a model-in-the-loop approach that iteratively probes and improves models. \newcite{Kaushik2019learningdifference} offer a causal account of spurious patterns, and counterfactually augment NLI datasets by editing examples to break the model. That approach is human-in-the-loop, using humans to find problems with one single model. In this work, we employ both human and model-based strategies iteratively, in a form of human-and-model-in-the-loop training, to create completely \emph{new} examples, in a potentially never-ending loop~\cite{Mitchell2018nell}.

Human-and-model-in-the-loop training is not a new idea. Mechanical Turker Descent proposes a gamified environment for the collaborative training of grounded language learning agents over multiple rounds \cite{Yang2017mastering}. The ``Build it Break it Fix it'' strategy in the security domain~\cite{Ruef2016buildit} has been adapted to NLP \cite{Ettinger2017towards} as well as dialogue safety \cite{Dinan2019build}. The QApedia framework \cite{Kratzwald2019learning} continuously refines and updates its content repository using humans in the loop, while human feedback loops have been used to improve image captioning systems \cite{Ling2017teaching}.  \newcite{Wallace2019Trick} leverage trivia experts to create a model-driven adversarial question writing procedure and generate a small set of challenge questions that QA-models
fail on. Relatedly, \newcite{lan2017ppdb} propose a method for continuously growing a dataset of paraphrases.

There has been a flurry of work in constructing datasets with an adversarial component, such as 
Swag \cite{Zellers2018swag} and HellaSwag \cite{Zellers2019hellaswag}, CODAH \cite{Chen2019codah}, Adversarial SQuAD \cite{Jia2017adversarial}, Lambada \cite{Paperno2016lambada} and others. Our dataset is not to be confused with abductive NLI \cite{Bhagavatula2019abductive}, which calls itself $\alpha$NLI, or ART.


\vspace{-3pt}
\section{Discussion \& Conclusion}
\vspace{-3pt}
In this work, we used a human-and-model-in-the-loop training method to collect a new benchmark for natural language understanding. The benchmark is designed to be challenging to current state-of-the-art models. Annotators were employed to act as adversaries, and encouraged to find vulnerabilities that fool the model into misclassifying, but that another person would correctly classify. We found that non-expert annotators, in this gamified setting and with appropriate incentives, are remarkably creative at finding and exploiting weaknesses. We collected three rounds, and as the rounds progressed, the models became more robust and the test sets for each round became more difficult. Training on this new data yielded the state of the art on existing NLI benchmarks.

The ANLI benchmark presents a new challenge to the community. It was carefully constructed to mitigate issues with previous datasets, and was designed from first principles to last longer. 
The dataset also presents many opportunities for further study. For instance, we collected annotator-provided explanations for each example that the model got wrong. We provided inference labels for the development set, opening up possibilities for interesting more fine-grained studies of NLI model performance. While we verified the development and test examples, we did not verify the correctness of each training example, which means there is probably some room for improvement there.

A concern might be that the static approach is probably cheaper, since dynamic adversarial data collection requires a verification step to ensure examples are correct. However, verifying examples is probably also a good idea in the static case, and adversarially collected examples can still prove useful even if they didn't fool the model and weren't verified. Moreover, annotators were better incentivized to do a good job in the adversarial setting. Our finding that adversarial data is more data-efficient corroborates this theory. Future work could explore a detailed cost and time trade-off between adversarial and static collection.

It is important to note that our approach is model-agnostic. HAMLET was applied against an ensemble of models in rounds 2 and 3, and it would be straightforward to put more diverse ensembles in the loop to examine what happens when annotators are confronted with a wider variety of architectures.

The proposed procedure can be extended to other classification tasks, as well as to ranking with hard negatives either generated (by adversarial models) or retrieved and verified by humans. It is less clear how the method can be applied in generative cases.

Adversarial NLI is meant to be a challenge for measuring NLU progress, even for as yet undiscovered models and architectures. 
Luckily, if the benchmark does turn out to saturate quickly, we will always be able to collect a new round.

\vspace{-5pt}
\section*{Acknowledgments}
\vspace{-5pt}
YN interned at Facebook. YN and MB were sponsored by DARPA MCS Grant \#N66001-19-2-4031, ONR Grant \#N00014-18-1-2871, and DARPA YFA17-D17AP00022. Special thanks to Sam Bowman for comments on an earlier draft.

\bibliography{acl2020}
\bibliographystyle{acl_natbib}

\clearpage
\appendix
\input{supplementary/realappendix.tex}

\end{document}

%% file: supplementary/realappendix.tex
\section{\label{appendix:stresstests}Performance on challenge datasets}


\begin{table*}[t]
\centering
\small
	\begin{tabular}{lcccccccccccccccc}
		\toprule
		 & \multicolumn{7}{c}{\bf Other Datasets } & \multicolumn{6}{c}{\bf ANLI}
		  \\ 
		 & \multicolumn{2}{c}{\bf SNLI} & \multicolumn{2}{c}{\bf MNLI$_{m}$} & \multicolumn{2}{c}{\bf MNLI$_{mm}$} & \bf F &  \multicolumn{2}{c}{\bf A1} & \multicolumn{2}{c}{\bf A2 } &  \multicolumn{2}{c}{\bf A3}  \\
		\bf Tag & \% c & \% h & \% c & \% h & \% c & \% h & \% claim & \% c &  \% h &  \% c &  \% h &  \% c  & \% h   \\
		\midrule
		Negation &$<1$ & 1& 14 & 16 & 12 &  16 & 3 & 2  & 6   &	3  & 10  &  22 & \it 14 \\
		`and'  & 30& 7 & 41 & 15 & 42 & 18 & 6 & 85  & \it 12 & 88 & 11 & 75 & 11  \\
		`or' & 1 & $<1$ & 7 & 2 & 8 & 2 & $<1$ &  6   & 0   & 	6  & $<1$ & 15 & 1 \\
	    Numbers & 10 & 4 & 16 & 8 & 15 & 9 & 9 & 72  & \bf 30  & \bf 73 & \bf 27 & \bf 42 & \bf 15 \\
	    Time   & 12 & 4 & 15 & 7 & 16 & 9 & 6 & 57  & \bf 22 & \bf 56 & \bf 19  & \bf 49 & \bf 11\\
		WH-words & 3& 1 & 16 & 7 & 18 & 9 & 2 &  28  & \it 5  & 27 & \it 5  & 35 & \it 5 \\
		Pronouns  & 11 & 7 & 37 & 20 & 39 & 24 & 2 &  30   & 9   & 28 & 7 &  60 & 13 \\
		Quantifiers  & 5& 3 & 21 & 16 & 22 & 17 & 3 & 14   & 10  & 17  & \it 12 & 38 & \it 12  \\
		Modals &$<1$ & $<1$& 17 &  13 & 18 &  14 & $<1$ &  2   & 3  & 3  & 2  & 35 & \it 14 \\
		$>$20 words  & 14 & $<1$ & 37 &  2  & 39 &  3 & $<1$ & 100 & \it 5 & 100 & \it 4 & 98 & \it 4 \\ \midrule
		\# exs & \multicolumn{2}{c}{ 10k} & \multicolumn{2}{c}{ 10k} &  \multicolumn{2}{c}{ 10k} & 9999 & \multicolumn{2}{c}{ 1k}&  \multicolumn{2}{c}{ 1k} & \multicolumn{2}{c}{ 1200}  \\ 
		\bottomrule
	\end{tabular}
	\caption{\label{tab:autotag}Percentage of development set sentences with tags in several datasets: AdvNLI, SNLI, MuliNLI and FEVER. `\%c' refers to percentage in contexts, and`\%h' refers to percentage in hypotheses. Bolded values label linguistic phenomena that have higher incidence in adversarially created hypotheses than in hypotheses from other NLI datasets, and italicized values have roughly the same (within 5\%) incidence.
	} 
\end{table*}

Recently, several hard test sets have been made available for revealing the biases NLI models learn from their training datasets \citep{nie2017,Mccoy2019right, Gururangan2018annotation,naik-EtAl:2018:C18-1}. We examine model performance on two of these: the SNLI-Hard \cite{Gururangan2018annotation} test set, which consists of examples that hypothesis-only models label incorrectly, and the NLI stress tests \citep{naik-EtAl:2018:C18-1}, in which sentences containing antonyms pairs, negations, high word overlap, i.a., are heuristically constructed. We test our models on these stress tests after tuning on each test's respective development set to account for potential domain mismatches. 
For comparison, we also report results from the original papers: for SNLI-Hard from \citeauthor{Gururangan2018annotation}'s implementation of the hierarchical tensor-based Densely Interactive Inference Network \citep[DIIN]{gong2018} on MNLI, and for the NLI stress tests, \citeauthor{naik-EtAl:2018:C18-1}'s implementation of InferSent \citep{conneau2017} trained on SNLI. 




\section{\label{appendix:linguisticanalysis}Further linguistic analysis}

We compare the incidence of linguistic phenomena in ANLI with extant popular NLI datasets to get an idea of what our dataset contains. We observe that FEVER and SNLI datasets generally contain many fewer hard linguistic phenomena than MultiNLI and ANLI (see \autoref{tab:autotag}).

ANLI and MultiNLI have roughly the same percentage of hypotheses that exceeding twenty words in length, and/or contain negation (e.g., `never', 'no'), tokens of `or', and modals (e.g., `must', `can'). MultiNLI hypotheses generally contains more pronouns, quantifiers (e.g., `many', `every'), WH-words (e.g., `who', `why'), and tokens of `and' than do their ANLI counterparts---although A3 reaches nearly the same percentage as MultiNLI for negation, and modals. However, ANLI contains more cardinal numerals and time terms (such as `before', `month', and `tomorrow') than MultiNLI. These differences might be due to the fact that the two datasets are constructed from different genres of text. Since A1 and A2 contexts are constructed from a single Wikipedia data source (i.e., HotPotQA data), and most Wikipedia articles include dates in the first line, annotators appear to prefer constructing hypotheses that highlight numerals and time terms, leading to their high incidence.

Focusing on ANLI more specifically, A1 has roughly the same incidence of most tags as A2 (i.e., within 2\% of each other), which, again, accords with the fact that we used the same Wikipedia data source for A1 and A2 contexts. A3, however, has the  highest incidence of every tag (except for numbers and time) in the ANLI dataset. This could be due to our sampling of A3 contexts from a wider range of genres, which likely affected how annotators chose to construct A3 hypotheses; this idea is supported by the fact that A3 contexts differ in tag percentage from A1 and A2 contexts as well. The higher incidence of all tags in A3 is also interesting, because it could be taken as providing yet another piece of evidence that our HAMLET data collection procedure generates increasingly more difficult data as rounds progress.

\begin{table*}[t]
    \centering
    \small 
    \begin{tabular}{lccc}
        \toprule
        & \multicolumn{3}{c}{\textbf{Entailment / Neutral / Contradiction}}\\
        \textbf{Round} & Train & Dev & Test\\\midrule
        A1 & 5,371 / 7,052 / 4,523 & 334 / 333 / 333 & 334 / 333 / 333 \\ 
        A2 & 14,448 / 20,959 / 10,053 & 334 / 333 / 333 & 334 / 333 / 333 \\ 
        A3 & 32,292 / 40,778 / 27,389 & 402 / 402 / 396 & 402 / 402 / 396 \\ \midrule  
        ANLI & 52,111 / 68,789 / 41,965 & 1,070 / 1,068 / 1,062  & 1,070 / 1,068 /1,062  \\ 
        \bottomrule
    \end{tabular}
    \caption{\label{tab:label_dist} Label distribution in splits across rounds.}
\end{table*}

\begin{table}[t]
    \centering
    \small
    \begin{tabular}{lrrr}
    \toprule
        Round & Dev + Test & Dev & Test \\\midrule
        A1 & 0.7210 & 0.7020 & 0.7400 \\
        A2 & 0.6910 & 0.7100 & 0.6720 \\
        A3 & 0.6786 & 0.6739 & 0.6832 \\
        \bottomrule
    \end{tabular}
    \caption{Inter-annotator agreement (Fleiss' kappa) for writers and the first two verifiers.}
    \label{tab:iaa-fleiss}
\end{table}

\begin{table}[t]
    \centering
    \small
    \begin{tabular}{rrrrr}
    \toprule
        SNLI & MNLI & A1 & A2 & A3\\\midrule
        85.8 & 85.2 & 86.1 & 84.6 & 83.9\\\bottomrule
    \end{tabular}
    \caption{Percentage of agreement of verifiers (``validators'' for SNLI and MNLI) with the author label.}
    \label{tab:iaa-agreement}
\end{table}

\section{\label{appendix:datasetproperties}Dataset properties}

\begin{figure*}[p]
	\centering
    \includegraphics[clip,width=0.95\textwidth]{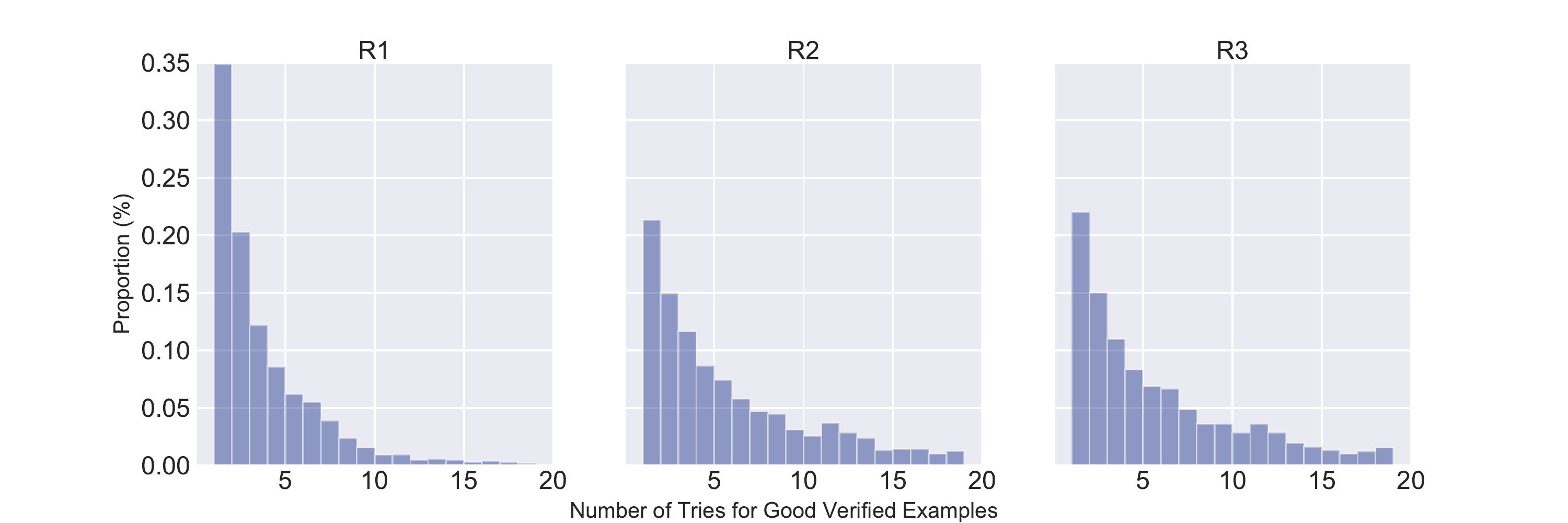}
 \caption{\label{fig:num_of_tries_good_verified}Histogram of the number of tries for each good verified example across three rounds.}
\end{figure*}

\begin{figure*}[p]
	\centering
    \includegraphics[clip,width=0.95\textwidth]{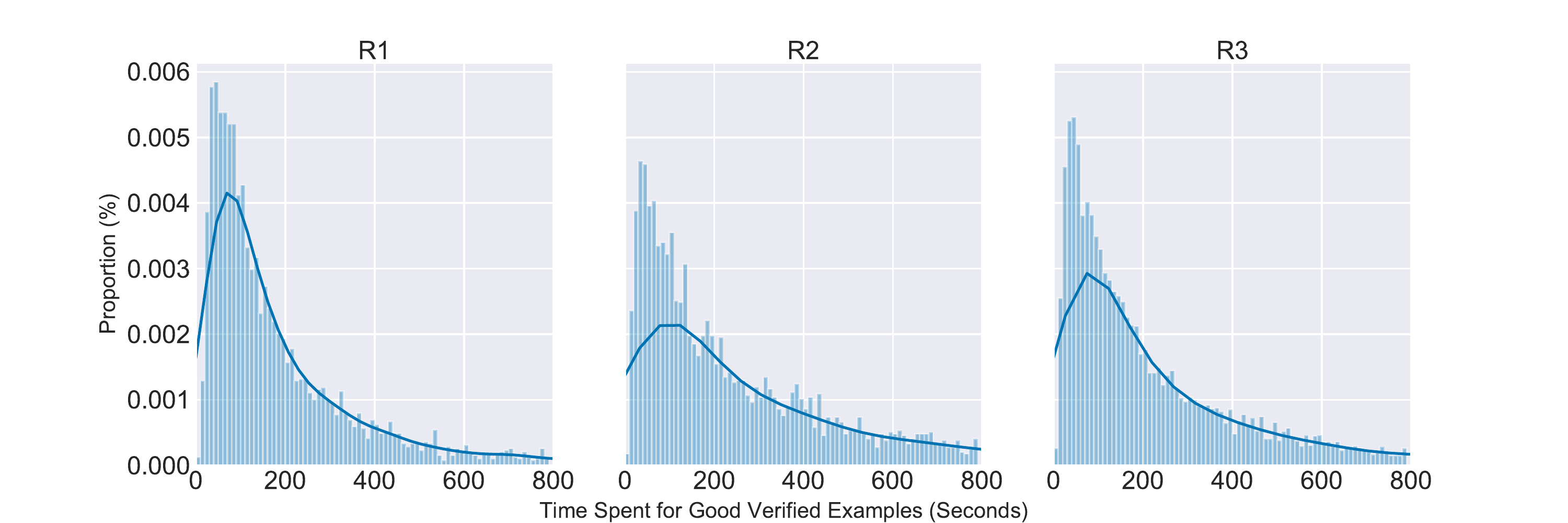}
 \caption{\label{fig:time_per_example}Histogram of the time spent per good verified example across three rounds.}
\end{figure*}

\begin{figure*}[p]
	\centering
    \includegraphics[clip,width=0.95\textwidth]{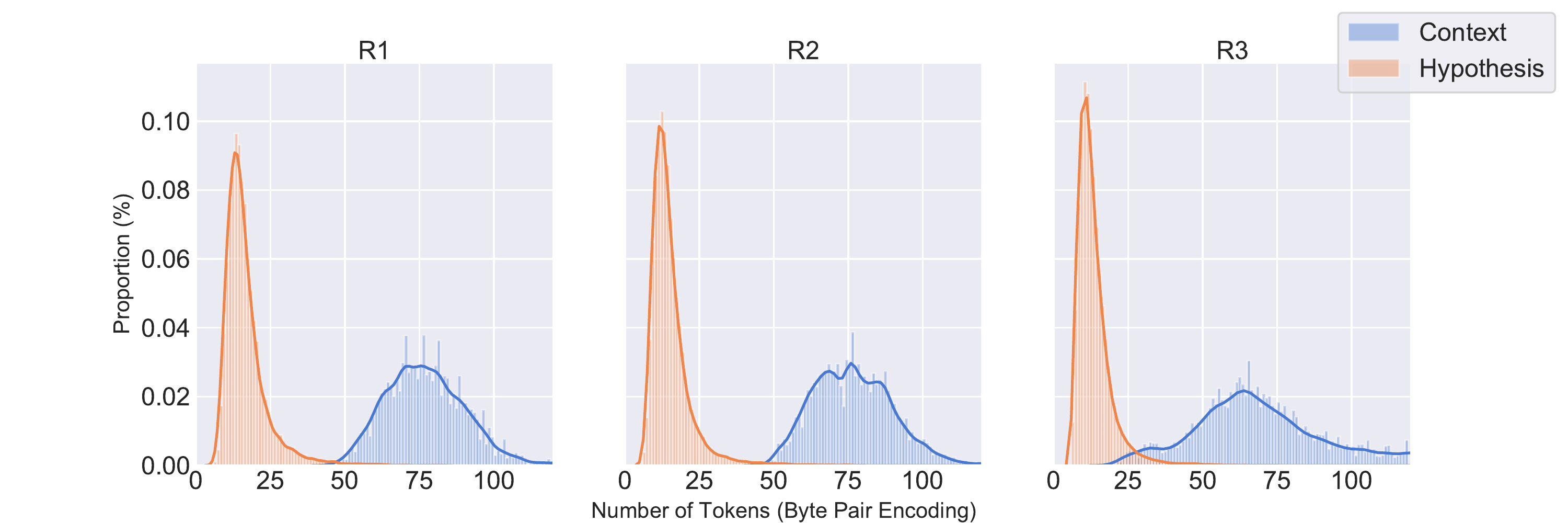}
 \caption{\label{fig:num_of_tokens}Histogram of the number of tokens in contexts and hypotheses across three rounds.}
\end{figure*}

\begin{figure*}[p]
	\centering
    \includegraphics[clip,width=0.95\textwidth]{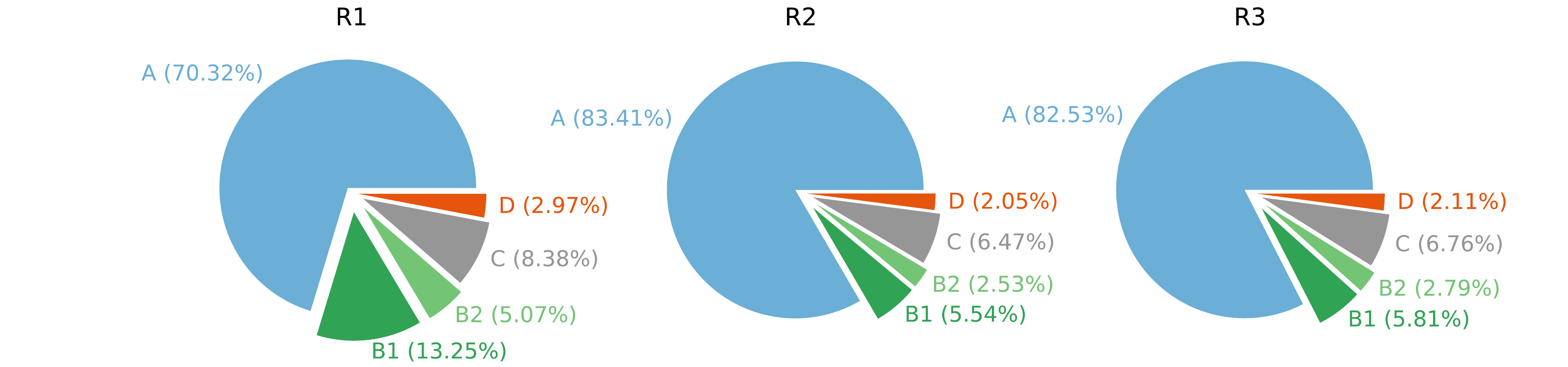}
 \caption{\label{fig:example_proportion}Proportion across three rounds. \textbf{A}=Examples that model got right, \textbf{B1}=Examples that model got wrong and the first two verifiers agreed with the writer, \textbf{B2}=Examples that model got wrong and only one of the first two verifiers agreed with the writer and a third verifier also agreed with the writer, \textbf{C}=Examples where two verifiers agreed with each other and overruled the writer, \textbf{D}=Examples for which there is no agreement among verifiers. \textbf{A} and \textbf{C} are added only to training set. \textbf{B1} and \textbf{B2} are added to training, dev, or test set. \textbf{D} was discarded.
}
\end{figure*}

Table \ref{tab:label_dist} shows the label distribution. Figure \ref{fig:num_of_tries_good_verified} shows a histogram of the number of tries per good verified example across for the three different rounds. Figure \ref{fig:time_per_example} shows the time taken per good verified example. Figure \ref{fig:num_of_tokens} shows a histogram of the number of tokens for contexts and hypotheses across three rounds. Figure~\ref{fig:example_proportion} shows the proportion of different types of collected examples across three rounds.

\paragraph{Inter-annotator agreement}

Table \ref{tab:iaa-fleiss} reports the inter-annotator agreement for verifiers on the dev and test sets. For reference, the Fleiss' kappa of FEVER \cite{Thorne2018fever} is $0.68$ and of SNLI \cite{Bowman2015snli} is $0.70$. Table \ref{tab:iaa-agreement} shows the percentage of agreement of verifiers with the intended author label.

\begin{table*}[ht]
\centering
\tiny
	\begin{tabular}{p{19em}p{9em}p{14em}p{4em}cccp{7.7em}}
		\toprule
        \multirow{2}{*}{\bf Context} &  \multirow{2}{*}{\bf Hypothesis} &  \multirow{2}{*}{\bf Reason} &
        \multirow{2}{*}{\bf Round} & \multicolumn{3}{c}{\bf Labels} & \multirow{2}{*}{\bf Annotations} \\
         & & & & orig. & pred. & valid. & \\
		\midrule
		Eduard Schulte (4 January 1891 in D\"usseldorf – 6 January 1966 in Z\"urich) was a prominent German industrialist. He was one of the first to warn the Allies and tell the world of the Holocaust and systematic exterminations of Jews in Nazi Germany occupied Europe. & Eduard Schulte is the only person to warn the Allies of the atrocities of the Nazis. & The context states that he is not the only person to warn the Allies about the atrocities committed by the Nazis. & A1 (Wiki) & C & N & C C & Tricky Presupposition, Numerical Ordinal\\ \midrule
		Kota Ramakrishna Karanth (born May 1, 1894) was an Indian lawyer and politician who served as the Minister of Land Revenue for the Madras Presidency from March 1, 1946 to March 23, 1947. He was the elder brother of noted Kannada novelist K. Shivarama Karanth. &  Kota Ramakrishna Karanth has a brother who was a novelist and a politician & Although Kota Ramakrishna Karanth's brother is a novelist, we do not know if the brother is also a politician & A1 (Wiki) & N & E & N E N & Standard Conjunction, Reasoning Plausibility Likely, Tricky Syntactic\\ \midrule
		The Macquarie University Hospital (abbreviated MUH) is a private teaching hospital. Macquarie University Hospital, together with the Faculty of Medicine and Health Science, Macquarie University, formerly known as ASAM, Australian School of Advanced Medicine, will integrate the three essential components of an academic health science centre: clinical care, education and research. & The Macquarie University Hospital have still not integrated the three essential components of an academic health science centre: clinical care, education and research & the statement says that the universities are getting together but have not integrated the systems yet & A1 (Wiki) & E & C & E E & Tricky Presupposition, Standard Negation \\ \midrule
	  Bernardo Provenzano (31 January 1933 -- 13 July 2016) was a member  of the Sicilian Mafia (``Cosa Nostra'') and  was suspected of having   been the head of the Corleonesi, a Mafia faction that originated in 	    the town  of Corleone, and de facto ``capo di tutti capi'' (boss of  all bosses) of the entire Sicilian Mafia until his arrest in 2006. 	     & It was never confirmed that Bernardo Provenzano was the leader of the Corleonesi. & Provenzano was only suspected as the leader of the mafia. It wasn't confirmed. &  A2 (Wiki) & E & N & E E & Tricky Presupposition, Standard Negation\\ \midrule
	  HMAS ``Lonsdale'' is a former Royal Australian Navy (RAN) training base that was located at Beach Street, Port Melbourne , Victoria, Australia. Originally named ``Cerberus III'', the Naval Reserve Base was commissioned as HMAS ``Lonsdale'' on 1 August 1940 during the Second World War. & Prior to being renamed, Lonsdale was located in Perth, Australia. & A naval base cannot be moved - based on the information in the scenario, the base has always been located in Victoria. & A2 & C & N & C C & Tricky Presupposition, Reasoning Facts \\ \midrule
	  Toolbox Murders is a 2004 horror film directed by Tobe Hooper, and written by Jace Anderson and Adam Gierasch. It is a remake of the 1978 film of the same name and was produced by the same people behind the original. The film centralizes on the occupants of an apartment who are stalked and murdered by a masked killer. & Toolbox Murders is both 41 years old and 15 years old. & Both films are named Toolbox Murders one was made in 1978, one in 2004. Since it is 2019 that would make the first 41 years old and the remake 15 years old. & A2 (Wiki) & E & C & E E & Reasoning Facts, Numerical Cardinal Age, Tricky Wordplay\\ \midrule
	  A biker is critically ill in hospital after colliding with a lamppost in Pete The incident happened at 1.50pm yesterday in Thorpe Road. The 23-year-old was riding a Lexmoto Arrow 125 when, for an unknown reason, he left the road and collided with a lamppost. He was taken to James Cook University Hospital, in Middlesbrough, where he remains in a critical condition. Any witnesses to the collision are asked to call Durham Police on 101, quoting incident number 288 of July 9. & The Lamppost was stationary. & Lampposts don't typically move. & A3 (News) & E & N & E E & Reasoning Facts, Standard\\	\midrule
	  ``We had to make a decision between making payroll or paying the debt,'' Melton said Monday. ``If we are unable to make payroll Oct. 19, we will definitely be able to make it next week Oct. 26 based on the nature of our sales taxes coming in at the end of the month. However we will have payroll the following week again on Nov. 2 and we are not sure we will be able to make that payroll because of the lack of revenue that is coming in.'' & The company will not be able to make payroll on October 19$^{th}$ and will instead dispense it on October 26$^{th}$ & It's not definitely correct nor definitely incorrect because the company said ``if'' they can't make it on the 19$^{th}$ they will do it on the 26$^{th}$, they didn't definitely say they won't make it on the 19$^{th}$ & A3 (News) & N & E & N C N & Reasoning Plausibility Likely, Tricky Presupposition\\ \midrule
	 The Survey: Greg was answering questions. He had been asked to take a survey about his living arrangements. He gave all the information he felt comfortable sharing. Greg hoped the survey would improve things around his apartment. THe complex had really gone downhill lately. & He gave some of the information he felt comfortable sharing. & Greg gave all of the information he felt comfortable, not some.  It was difficult for the system because it couldn't tell a significant difference between to word ``some'' and ``all.'' & A3 (Fiction) & C & E & C C & Tricky (Scalar Implicature) \\
	  \bottomrule
	\end{tabular}
	\caption{\label{tab:moreexamples} Extra examples from development sets. `A$n$' refers to round number, `orig.' is the original annotator's gold label, `pred.' is the model prediction, `valid.' is the validator labels, `reason' was provided by the original annotator, `Annotations' is the tags determined by linguist expert annotator.
	} 
\end{table*}
\section{\label{appendix:moreexamples}Examples}

We include more examples of collected data in Table \ref{tab:moreexamples}.

\section{\label{appendix:userinterface}User interface}

\begin{figure*}[p]
	\centering
    \includegraphics[clip,width=0.95\textwidth]{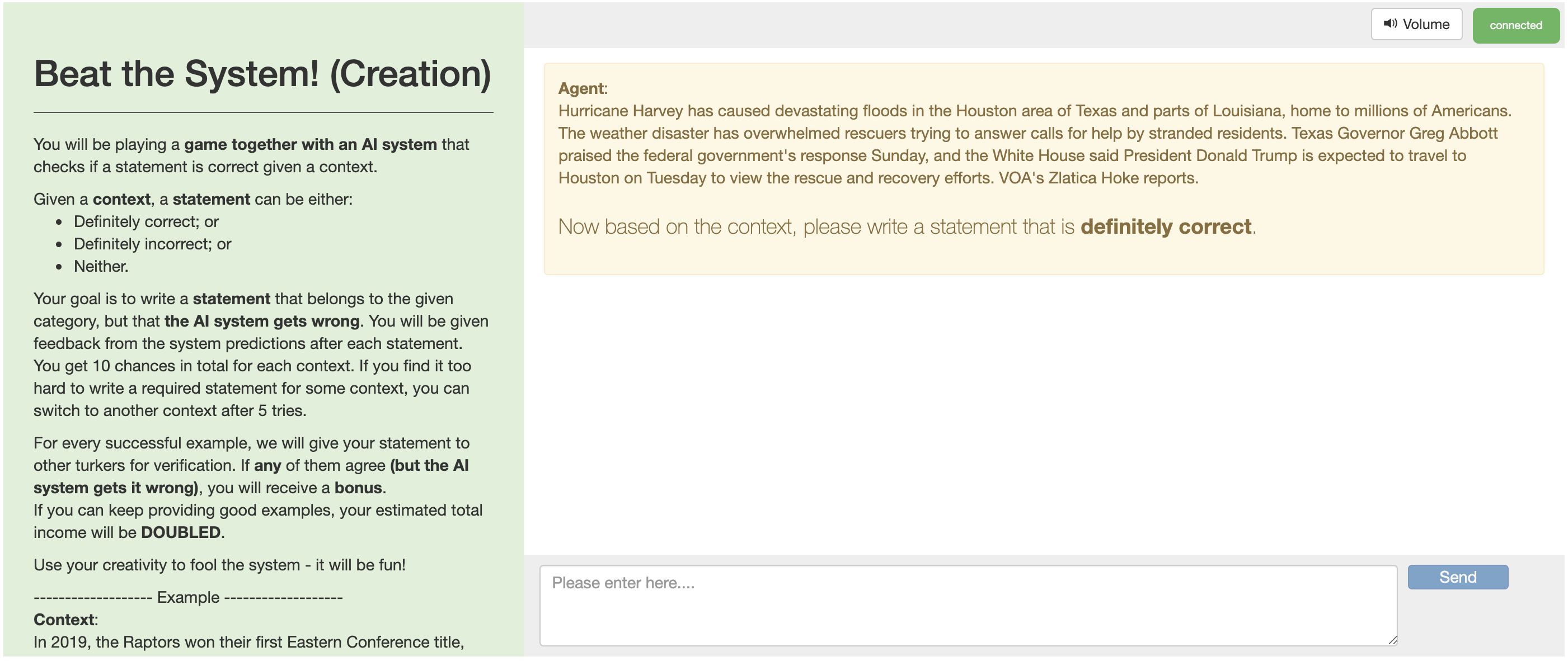}
 \caption{\label{fig:parlai_interface_context}UI for Creation. (Provide the context to annotator)}
\end{figure*}

\begin{figure*}[p]
	\centering
    \includegraphics[clip,width=0.95\textwidth]{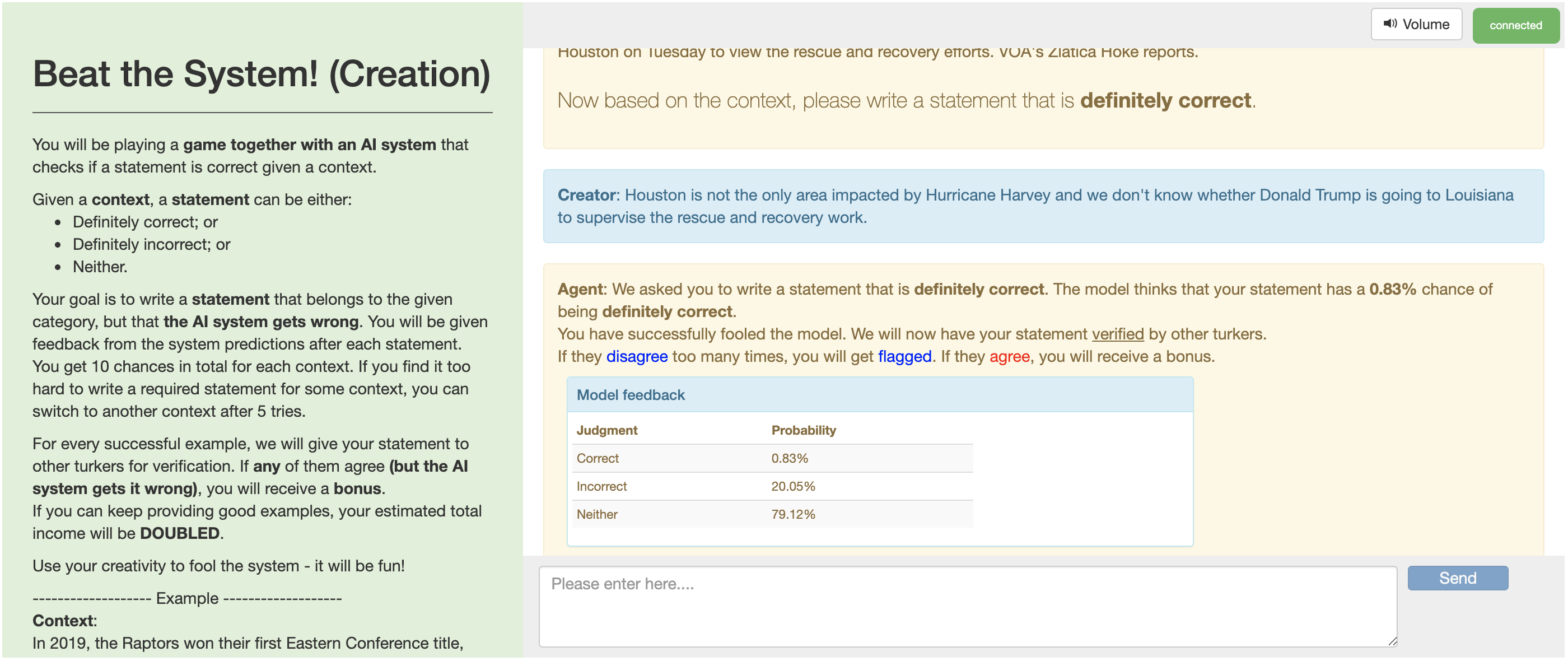}
 \caption{\label{fig:parlai_interface_feedback}
 Collection UI for Creation. (Give the model feedback to annotator)}
\end{figure*}

\begin{figure*}[p]
	\centering
    \includegraphics[clip,width=0.95\textwidth]{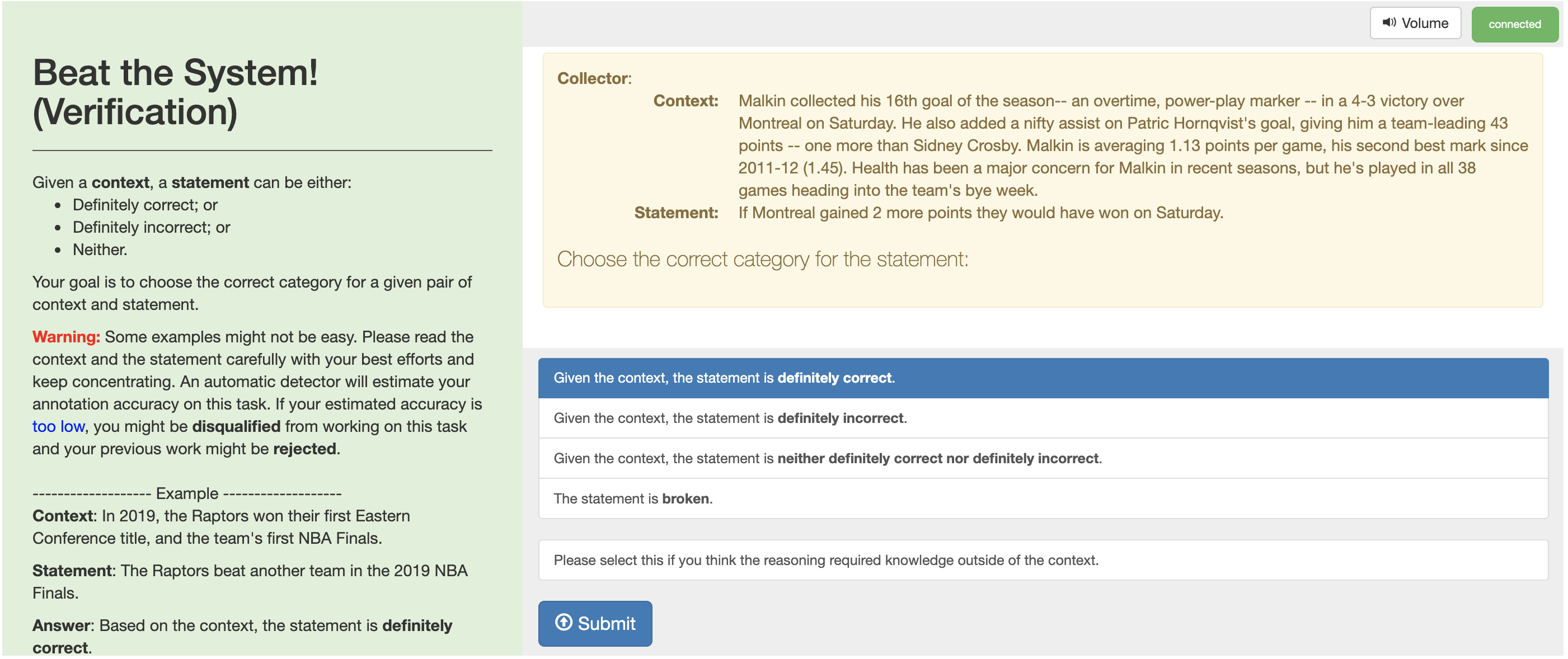}
 \caption{\label{fig:parlai_interface_verification}UI for Verification Task.}
\end{figure*}

Examples of the user interface are shown in Figures \ref{fig:parlai_interface_context}, \ref{fig:parlai_interface_feedback} and \ref{fig:parlai_interface_verification}.


